\documentclass[lettersize,journal]{IEEEtran}
\usepackage{amsmath,amsfonts}
\usepackage{algorithmic}
\usepackage{array}
\usepackage[caption=false,font=normalsize,labelfont=sf,textfont=sf]{subfig}
\usepackage{textcomp}
\usepackage{stfloats}
\usepackage{url}
\usepackage{verbatim}
\usepackage{graphicx}
\def\BibTeX{{\rm B\kern-.05em{\sc i\kern-.025em b}\kern-.08em
    T\kern-.1667em\lower.7ex\hbox{E}\kern-.125emX}}
\usepackage{balance}

\usepackage{graphicx}  

\DeclareGraphicsExtensions{.pdf,.png,.jpg,.jpeg} 

\graphicspath{{figures/}{pictures/}{images/}{./}} 

\usepackage{microtype}                 
\PassOptionsToPackage{warn}{textcomp}  
\usepackage{textcomp}                  
\usepackage{mathptmx}                  

\usepackage{enumitem}

\usepackage{times}                     
\usepackage{cite}                      
\usepackage{tabu}                      
\usepackage{booktabs}                  

\usepackage{amsmath} 
\usepackage{amssymb} 
\usepackage{amsthm}
\usepackage{bm}
\usepackage[ruled,vlined]{algorithm2e}
\usepackage{subfiles}
\usepackage{hyperref}
\usepackage{tikz}
\usepackage{pgfplots}
\usepackage{placeins}
\usepackage[caption=false,font=normalsize,labelfont=sf,textfont=sf]{subfig}
\pgfplotsset{compat=1.17}
\usepgfplotslibrary{groupplots}
\pgfplotsset{
  every axis/.append style={
    no markers,
    grid=major,
    grid style={dashed},
    legend style={font=\tiny},
    ylabel style={font=\scriptsize},
    xlabel style={font=\scriptsize},
    width=\linewidth
  },
  every axis plot/.append style={line width=1.2pt, line join=round},
  every axis legend/.append style={legend columns=2},
  group/group size=3 by 1,
  every x tick label/.append style={alias=XTick,inner xsep=0pt},
  every x tick scale label/.style={at=(XTick.base east),anchor=base west}
}

\definecolor{col1}{RGB}{36, 61, 93}
\definecolor{col2}{RGB}{77, 143, 145}
\definecolor{col3}{RGB}{228, 206, 135}
\definecolor{col4}{RGB}{186, 161, 22}
\definecolor{col5}{RGB}{45, 3, 59}
\definecolor{col6}{RGB}{193, 71, 233}
\definecolor{col7}{RGB}{126, 134, 255}
\definecolor{col8}{RGB}{85, 170, 255}

\tikzset{
  curve1/.style={col1},
  curve2/.style={col2},
  curve3/.style={col3},
  curve4/.style={col4},
  curve5/.style={col5},
  curve6/.style={col6},
  curve7/.style={col7},
  curve8/.style={col8},
  curve9/.style={col1, densely dotted},
  curve10/.style={col2, dotted},
  curve11/.style={col3, dashdotted},
  curve12/.style={col4, dashed},
}

\DeclareMathOperator*{\argmin}{arg\,min}

\newtheoremstyle{theoremstyle} 
    {\topsep}                    
    {\topsep}                    
    {}                   
    {}                           
    {\bfseries}                   
    { :}                          
    {.5em}                       
    {}  
\theoremstyle{theoremstyle}

\newtheorem{Hyp}{Assumption}
\newtheorem{Rk}{Remark}
\newtheorem{Cond}{Conditions}

\newcommand{\step}[2]{\textrm{---} \textit{Step #1}: #2}

\newcommand{\br}{\mathfrak{b}}
\newcommand{\M}{\mathcal{M}}

\newcommand{\N}{\mathbb{N}}
\newcommand{\R}{\mathbb{R}}

\newcommand{\Dict}{\mathcal{D}}

\newcommand{\Lamb}{\bm{\lambda}}

\newcommand{\keanu}[1]{\textcolor{black}{#1}}

\begin{document}
\title{Robust Barycenters of Persistence Diagrams}
\author{Keanu Sisouk, Eloi Tanguy, Julie Delon and Julien Tierny}

\maketitle

\begin{abstract}
This
{short}
paper presents
{a general approach}
for computing robust Wasserstein barycenters
\cite{agueh2011, Turner2014, vidal20} of persistence diagrams.
{The classical method}
consists in computing
{assignment arithmetic means}
after finding the
optimal transport plans between the barycenter and the persistence diagrams.
However, this
{procedure}
only works for
{{the} transportation cost} {related to}
the {$q$-}Wasserstein distance
{$W_q$ when $q=2$.}
We
{adapt an alternative}
fixed-point method
\cite{tanguy2025computingbarycentresmeasuresgeneric}
to compute a barycenter diagram for 
generic
transportation costs {($q > 1$)}, in
particular
those robust to outliers{, $q \in (1,2)$}. We 
{show}
the
utility of
{our work}
in two
applications{: \emph{(i)} the clustering of persistence diagrams on their
metric space and \emph{(ii)} the dictionary
encoding of 
persistence diagrams \cite{sisouk2024}. In both scenarios, we demonstrate the
added robustness to outliers provided by our generalized framework. Our Python 
implementation is available at this address: 
\href{https://github.com/Keanu-Sisouk/RobustBarycenter}{https://github.com/Keanu-Sisouk/RobustBarycenter}.}
\end{abstract}


\section{Introduction}
\IEEEPARstart{W}{ith} measurement devices and numerical techniques being more and more precise, the resulting datasets have become more and more complex geometrically. This complexity hinders the users during any exploration and analysis to interpret them.
Those challenges motivate the conception of expressive, informative{,}
concise and simple data abstractions, encoding the main features and patterns
of interest of the data.

Topological Data Analysis (TDA) \cite{edelsbrunner09}  is a family of tools
developed to address this challenge. It aims to provide simple objects, called
topological representations, describing the main topological structures in a
{dataset}. TDA has become a staple for analyzing scalar data, its
efficiency and robustness being shown in numerous visualization tasks
\cite{heine16}. It has proven to be successful in many applications, several
examples include turbulent combustion \cite{bremer_tvcg11, gyulassy_ev14,
laney_vis06}, material sciences \cite{favelier16, gyulassy_vis07,
gyulassy_vis15, soler_ldav19}, nuclear energy \cite{beiNuclear16}, fluid 
dynamics \cite{kasten_tvcg11, nauleau_ldav22}, bioimaging 
\cite{beiBrain18, topoAngler, carr04}, chemistry \cite{harshChemistry, 
chemistry_vis14, Malgorzata19, olejniczak_pccp23} or astrophysics 
\cite{shivashankar2016felix, sousbie11} to name a few.

Among the different topological representations, such as the merge and contour trees \cite{tarasov98, carr00,
MaadasamyDN12, AcharyaN15, CarrWSA16, gueunet_tpds19}, the Reeb graph 
\cite{biasotti08, pascucci07, tierny_vis09, parsa12, DoraiswamyN13, 
gueunet_egpgv19}, or the Morse-Smale complex \cite{forman98, EdelsbrunnerHZ01,
EdelsbrunnerHNP03,
BremerEHP03, Defl15, robins_pami11, ShivashankarN12, gyulassy_vis18}, the
Persistence Diagram (\autoref{fig:PDExample}) as been prominently used and studied. It is a simple and concise topological representation, which encodes the main topological features of  data.

With
{the}
increase of
geometrical complexity discussed above, a new challenge has come
forth as users are confronted to the emergence of \textit{ensemble datasets}.
With these representations, a given phenomenon is not only described by a single
dataset, but by a \textit{collection} of datasets, called \textit{ensemble
members}. This results in the analysis of an ensemble where each element is a
topological representation, such as {a} persistence diagram{.}

However, this process moves the problem of analyzing an ensemble of scalar
fields to analyzing an ensemble of persistence diagrams. To address this
challenge{,}
the
question of finding a good \textit{representative} of such an ensemble has
emerged.
{An established answer to that question is the notion of}
\textit{Wasserstein
barycenter} \cite{agueh2011, Turner2014} of persistence diagrams, based on the
so called
{$q$-}Wasserstein distance \cite{villani2008optimal}. To compute such
barycenter{s}, several algorithms have been proposed {for $q=2$} 
\cite{Turner2014,
lacombe2018, vidal20}. This
barycenter emulates the behavior of
an \textit{average},
{and}
as such{, it is also sensitive to outliers.}

Works on \textit{Wasserstein medians}
counter this issue
\cite{turner2019,Carlier_2024}, showing their
stability
properties.
{In the context of probability measures, extensions}
to {generic transportation costs}
 have been
introduced recently
\cite{brizzi2024,
tanguy2025computingbarycentresmeasuresgeneric}.

This {short} paper proposes to
{leverage
generic transportation costs for
computing robust $q$-Wasserstein barycenters of}
ensembles of persistence diagrams{, with $q > 1$}.
{For this, we adapted a recent fixed-point method
\cite{tanguy2025computingbarycentresmeasuresgeneric} from generic probability
measures to persistence diagrams.}
As discussed
above,
{the resulting}
barycenters have the advantage
{of being}
more \textit{robust} to the
presence of
{outlier diagrams in the ensemble.}
Experiments on synthetic and
real-life data showcase this property
{for typical ensembles of persistence diagrams used previously in the
literature.}
We
{show}
the utility of
{our}
barycenters
by using them in a clustering
problem
{on the Wasserstein metric space}
and by using them for a Wasserstein
dictionary encoding problem \cite{sisouk2024}. We give details on this
\textit{robust {barycentric}} framework in
\autoref{sec:RobustBary}, then we show the utility of these robust barycenters
with experiments in \autoref{sec:result} with two applications in
\autoref{sec:kmeans} and \autoref{sec:PDDict}.

\subsection{Related work}

The literature related to our work can be classified into two main categories, reviewed in the following: \textit{(i)} ensemble visualization, and \textit{(ii)} topological methods for ensemble of persistence diagrams.

\noindent \textbf{(i) Ensemble visualization:}
{A common approach}
to model data variability is using ensemble datasets. In this
setting, the variability is modeled by a succession of empirical observations
(i.e{.,} the \textit{members} of the ensemble). Existing approaches
compute geometrical objects encoding the features of interest (such as level
sets and streamlines for example) for each member of the ensemble. Then, a
\textit{representative} of this ensemble of geometrical objects can be
estimated. In light of this, several techniques has been established. For
example, spaghetti plots \cite{spaghettiPlot} are used to study level-set
variability, especially for weather data \cite{Potter2009, Sanyal2010}.
Box-plots \cite{whitaker2013,
Mirzargar2014} are used to analyze the variability of contours and curves. In the case of flow ensemble, Hummel et al. \cite{Hummel2013} has proposed a Lagrangian framework for classification purposes. More precisely, clustering techniques have been studied to highlight the main trends in ensemble of streamlines \cite{Ferstl2016} 
and isocontours \cite{Ferstl2016b}. However, only few approaches applied those strategies to topological representations. Favelier et al. \cite{favelier_vis18} and 
Athawale et al. \cite{athawale_tvcg19} introduced techniques for analyzing the
geometrical variability of critical points and gradient separatrices
respectively.
{Ensemble layouts have been proposed for contour trees}
\cite{Kraus2010VisualizationOU, Wu2013ACT}.
However,
{the}
above techniques do not focus on the computation of a representative of an
ensemble of topological objects.

\noindent \textbf{(ii) Topological methods: } To find a \textit{representative}
of an ensemble of persistence diagrams, notions from optimal transport
\cite{monge81, Kantorovich} were adapted to persistence diagrams. A central
notion is the so-called \textit{Wasserstein} distance
\cite{santambrogio2015optimal}. The Wasserstein distance between persistence
diagrams \cite{edelsbrunner09} (\autoref{sec:WassDist}) has been studied by the
TDA community \cite{CohenSteinerEH05, Cohen-Steiner2007}. This distance is
computed by solving an assignment problem,
{for which}
exact \cite{munkres1984elements} and approximate
\cite{Bertsekas81,Kerber2016} implementation can be found in open-source
\cite{ttk19, flamary2021pot}. Using this distance, the \textit{Wasserstein}
barycenter is used to find a \textit{representative} diagram of an ensemble,
\cite{agueh2011}. Turner et al. \cite{Turner2014} first introduced an algorithm
for the computation of such a barycenter for persistence diagrams, along with
convergence results and theoretical properties of the persistence diagram space.
Lacombe et al. \cite{lacombe2018} proposed a method to compute a barycenter
based on the entropic formulation of optimal transport \cite{cuturi2013,
cuturi2014}. However, this method requires a vectorization of persistence
diagrams, which is not only subject to parameters, but
{which also challenges visual analysis and inspection.}
Indeed, in this case the features of interest cannot
be tracked by the users during the analysis. Vidal et al. \cite{vidal20}
proposed an approach allowing the tracking of the features{. This}
method is based on a progressive
{framework,}
which accelerates the computation
time compared to Turner et al.'s method \cite{Turner2014}. To take it further,
several authors have proposed methods to find {a} representation of
ensembles of topological descriptors by a basis of \textit{representative}
descriptors. Li et al. \cite{LI2023}
{leveraged}
sketching method{s}
\cite{woodRuff2014}
{for}
vectorize{d} merge trees. Pont et al.
\cite{pont2022principal} introduced a principal geodesic analysis method for
merge trees, and Sisouk et al. \cite{sisouk2024} brought forth a Wasserstein
dictionary encoding method for an ensemble of persistence diagrams. Both
{latter} methods avoid the difficulties associated with vectorizations
(e.g. quantization and linearization artifacts, inaccuracies in vectorization
reversal).
However,
{all of the above literature}
propose representatives that can be
sensible to the presence of outliers. Turner et al. \cite{turner2019} studied
the notion of median of a population of persistence diagrams, but no exact
algorithm nor computation was proposed. In this work, we
{describe}
a {general}
framework
{for}
computing a robust barycenter,
{by adapting a recent}
fixed point method
\cite{tanguy2025computingbarycentresmeasuresgeneric} {from generic
probability
measures to}
persistence diagrams. This robust
barycenter is more stable to the presence of outliers,
{thereby}
enhancing
other analysis framework{s} such as clustering algorithms {or
dictionary-based encodings}.

\subsection{Contributions}

This paper makes the following contributions:
\begin{enumerate}

\item \textit{A
{general}
framework for
robust barycenter{s} of
persistence diagrams:}
{By adapting a recent approach
\cite{tanguy2025computingbarycentresmeasuresgeneric} from
probability
measures to persistence diagrams, we show how
barycenter diagrams can be reliably estimated, for
{generic $W_q$ distances (\autoref{sec:RobustBary}),}
despite
outliers
(\autoref{sec:result}).}

\item \textit{An application to clustering:} We present an application
to clustering  (\autoref{sec:kmeans}), where
{our work yields an improved robustness to the}
outlier
diagrams
{that are naturally present in ensembles used previously in the
literature.}

\item \textit{An application to Wasserstein dictionary {encoding:}}
We present an application to
dictionary encoding {(\autoref{sec:PDDict}),}
where {the added robustness of our generalized barycenters is
demonstrated over standard barycenters.}

\item \textit{Implementation: } We provide a Python implementation
of
{our work}
that can be used for reproducibility purposes.

\end{enumerate}

\section{Preliminaries}
This section presents the {required}
theoretical foundations
{to our work.}
We introduce
persistence diagrams (\autoref{sec:PDs}) and the usual metric
used in topological data analysis (\autoref{sec:WassDist}).

\begin{figure}[h!]
\centering
 \includegraphics[width=\columnwidth]{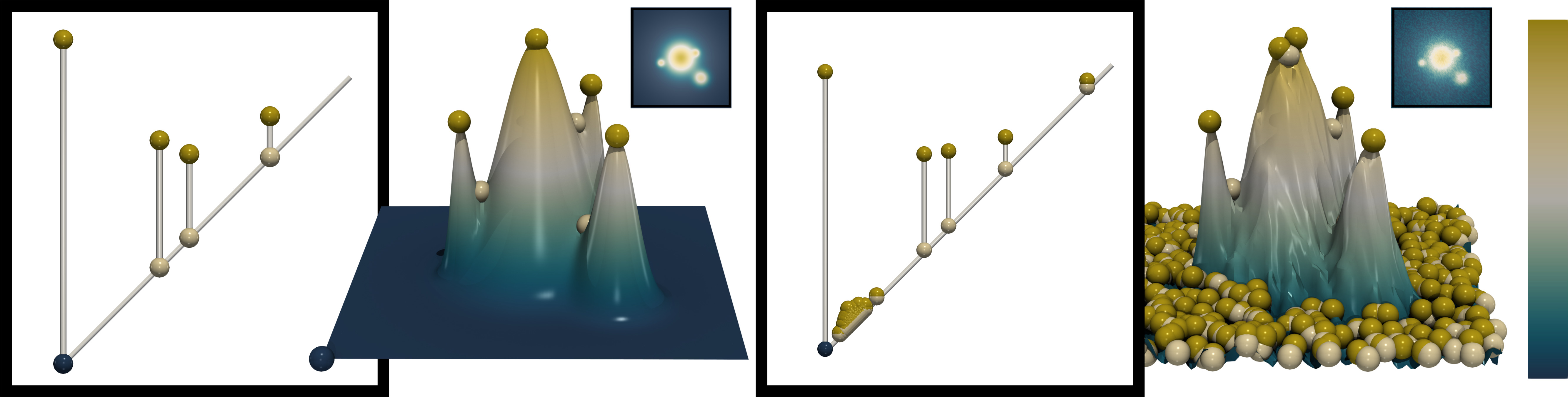}
\caption{Example of persistence diagrams of a
{smooth}
(left) and noisy scalar field (right). The four main features are
represented with long {bars}
in the persistence diagrams. In the noisy
diagram, the noise in the scalar field is encoded by small bars {near
the diagonal}.}
\label{fig:PDExample}
\end{figure}

\subsection{Persistence diagrams} \label{sec:PDs}

Given a scalar field $f$ on a $(d_{\M})$-manifold $\M$, with $d_{\M}
\leq 3$ {in our applications}, we denote $f^{-1}_{_\infty}(w) =
f^{-1}(]-\infty,w])$ the sub-level set of $f$ at value $w\in \R$.
While sweeping $w$ from $-\infty$ to $+\infty$, the topology of the set
$f^{-1}_{-\infty}(w)$ changes at specific values $w = f(c)${, associated
to the \emph{critical points} of $f$. These points}
are classified by their index
$\mathcal{I}$: 0 for minimal, 1 for $1$-saddles, $\hdots$, $d_{\M}-1$ for
$(d_{\M} -1)$-saddles, and $d_{\M}$ for maxima. The Elder rule
\cite{edelsbrunner09} states that each topological feature is associated to a
pair of
{critical points}
$(b,d)=  \big(f(c),f(c')\big)$, with $f(c) <
f(c')$, representing its \textit{birth} $b$ and \textit{death} $d$ in the sweep.
{These pairs can be visually represented as}
vertical
{bars in $\R^2$,} with horizontal coordinate $b$ and
vertical {coordinates $b$ and} $d${,} where $b-d$ encodes the
lifespan of the {associated} topological feature.
This representation is the so-called \textit{Persistence Diagram}.
{Formally, a}
persistence diagram is
the union of a finite set of 2D points $ X = \{x
= (b,d) \in \R^2 \ |\ b < d\}$, along with
the diagonal $\Delta =
\{(b,b)\ |\ b\in\R\}$ of $\R^2$ (the diagonal is not stored in any way in
practice). In most cases, important topological features stand out from the
diagonal while noise can be typically found near the diagonal
(\autoref{fig:PDExample}). In the following, we will simply denote a persistence
diagram by $X = \{x_1,\hdots, x_K\}$ with $K\in\N^*$.

\begin{figure}[tb]
\centering
 \includegraphics[width=\columnwidth]{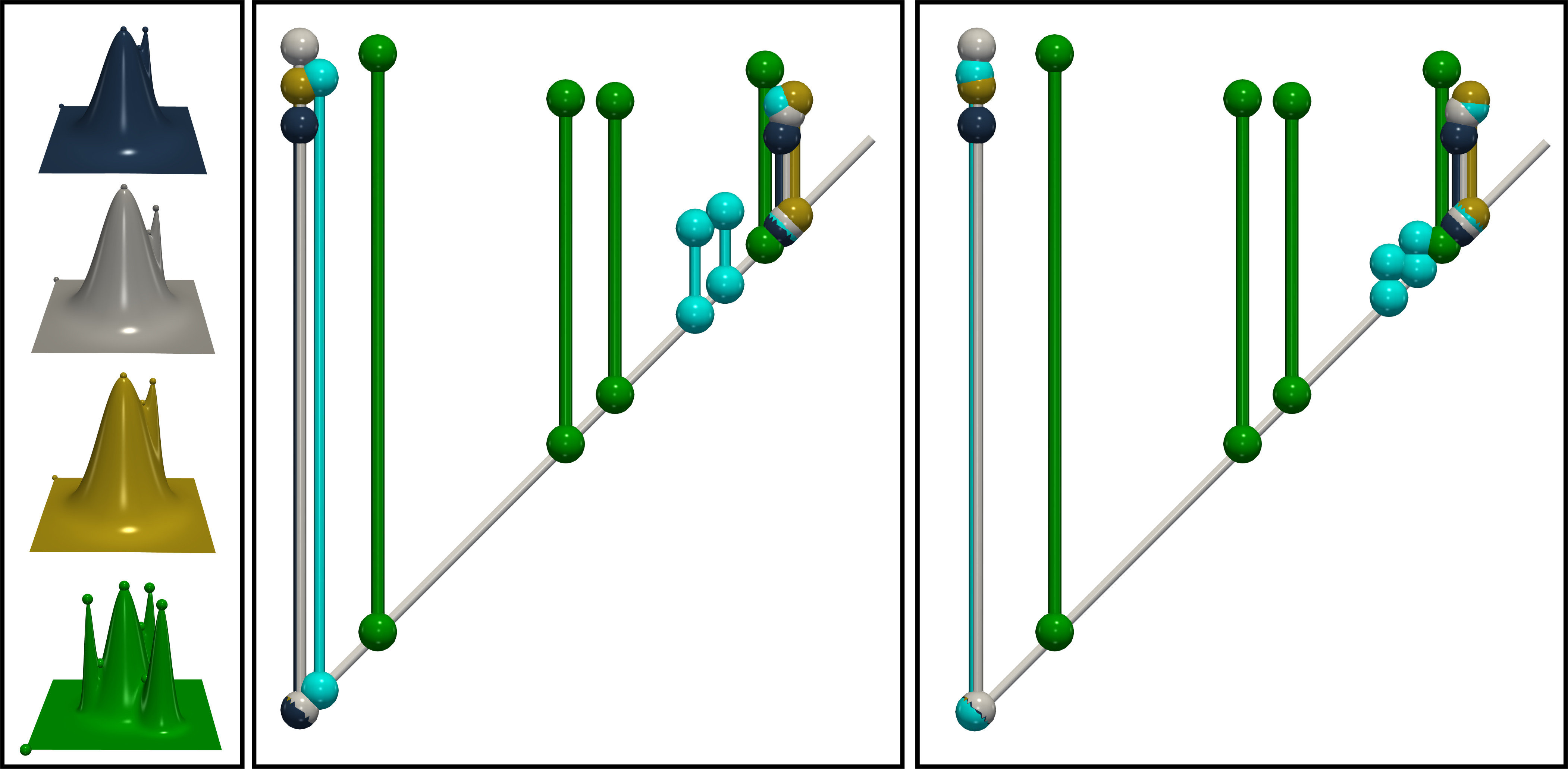}
\caption{Comparison of barycenters computed with different values of $q$. On the
left we have {terrain views of} four
scalar fields colored in
blue, gray, yellow and green{, the latter being an outlier}
{(featuring more peaks)}.  The corresponding
persistence diagrams {are represented with matching colors and the barycenters are represented in cyan.}
The barycenter with $q = 2$ {(center)} is more
sensitive to the presence of the {green} outlier,
{with two cyan bars of medium persistence, due to the outlier peaks
in the green dataset}. {For $q=1.5$ (right), the persistence of these two
bars is significantly reduced, and so will be their importance in distance
computations.}}
\label{fig:BaryComparison}
\end{figure}

\subsection{Wasserstein distance and barycenter} \label{sec:WassDist}

\noindent The Wasserstein distance is widely used
{to compare}
persistence diagrams. In topological data analysis, in order to use this
distance, a typical pre-processing involves augmenting the persistence diagrams.
Formally, let us consider $X = \{x_1,\hdots, x_{K_1}\}$ and $Y = \{y_1,\hdots,
y_{K_2}\}${,} two persistence diagrams. Taking a point $x =
(b,d)\in\R^2$, we denote by $\pi_{\Delta}(x) = \left(\frac{b+d}{2},
\frac{b+d}{2}\right)$ its projection on $\Delta$. We consider $\Delta_{X}$ and
$\Delta_Y$ the diagonal projections of the points of $X$ and $Y$ respectively.
We finish by taking $X' = X \cup \Delta_Y$ and $Y' = Y \cup \Delta_X${.
This}
results in $|X'| = |Y'| = K$. For the remainder of this paper, we consider that
{persistence diagrams are always augmented this way.}

\keanu{Then, for two persistence diagrams $X$ and $Y$, the $q$-Wasserstein distance $W_p$
is defined as:
\begin{equation}\label{eq:2Wass}
W_q(X,Y) = \displaystyle\min_{\phi: X\to Y} \left(
\sum\limits_{\ell = 1}^K c_q\big(x_{\ell},
\phi(x_{\ell})\big)\right)^{1/q},
\end{equation}
where $\phi: X\to Y$ is a bijection between $X$ and $Y$. The
{\emph{transportation cost}}
$c_q$, {based on powered distances,}} is such that $c_q(x,y) = \|x-y \|_2^q$
if $x\notin \Delta$ or $y\notin \Delta$, and 0 otherwise.
An optimal bijection $\phi^*$ minimizing \autoref{eq:2Wass} is called an optimal
transport plan.
{In practice, $q$ is often set to $2$, yielding the 2-Wasserstein
distance, noted $W_2$.} {For $q\geq 1$, the $q$-Wasserstein distance defines a metric on the space of persistence diagrams \cite{turner2019}.}
A Wasserstein {$W_q$}
barycenter of persistence diagrams $X_1,\hdots, X_m$\cite{Turner2014}, is defined by
{minimizing the Fréchet energy {over the space of persistence diagrams $\mathbb{D}$}:}
\begin{equation}\label{eq:2Bary}
\argmin_{{B\in\mathbb{D}}} \displaystyle\sum\limits_{i = 1}^m \lambda_i
W_{{q}}^{{q}}(B,X_i),
\end{equation}
where $\lambda_i \geq 0$ and $\sum_i\lambda_i = 1$ {(i.e $\Lamb = (\lambda_1,\hdots,\lambda_m)\in\Sigma_m$)}.
{Practical algorithms have been proposed for the computation
of Wasserstein barycenters \cite{Turner2014, vidal20}, but only for the
specific case where  $q = 2$. When $q = 2$, a}
solution $B^*$ of \autoref{eq:2Bary} has the following property: a point $x\in
B^*$ is an arithmetic mean of $m$ points each in $X_1,\hdots,X_m${,
which considerably eases the optimization of \autoref{eq:2Bary}}.
\autoref{fig:BaryComparison} {(center)} presents an example of a $W_2$
barycenter.
A barycenter computed using the $W_2$ distance emulates the
behavior of a mean of an ensemble of scalars. As such, it is prone to the
influence of outliers.
{This motivates the use of a more general framework}
for computing robust barycenter{s} (detailed in
\autoref{sec:RobustBary}).

\begin{algorithm}[h!]
\SetAlgoLined
\KwIn{ Set of persistence diagrams $\{X_1, \hdots , X_m\}$, barycentric weights $\lambda_1,\hdots,\lambda_m$, and iteration number $T$.}
\textbf{Output :} Wasserstein barycenter $B^{(T)} = \{x_1^{(T)},\hdots,x_K^{(T)}\}$.\

\textbf{Initialization: } $B^{(0)} = X_1$.\

\For{$  0\leq t \leq T-1$}{

		{\emph{// 1. Assignment step}}

	\For{$ i\in \{1,\hdots,m\}$}{
		{Compute $\phi_i \gets \displaystyle\argmin_{\phi:B^{(t)} \to
X_i}\sum\limits_{\ell=1}^K
		c_{q}\big(x_{\ell}^{(t)},\phi(x_{\ell}^{(t)})\big)$.}
	}\

	{\emph{// 2. Update step}}

	\For{$\ell\in\{1,\hdots,K\}$}{

		{\emph{// Ground barycenter computation}}

		Find $x_{\ell}^{(t+1)} = \br_{{q}}\big(\phi_1(x_{\ell}^{(t)}),\hdots,
\phi_m(x_{\ell}^{(t)})\big)$.
	}
	\If{$E_F^{(t)}-E_F^{(t+1)} < 10^{-3}$}{
	Break.
	}
}
 \caption{{Barycenter computation algorithm.}}
 \label{alg:WassBary}
\end{algorithm}
\section{Robust barycenter} \label{sec:RobustBary}
This section presents a
{general}
framework for computing barycenters of persistence
diagrams using $W_q$ distances, for arbitrary
$q$ values
such that $q > 1$.

\subsection{Optimization}
\label{sec_optimization}

{The optimization of \autoref{eq:2Bary} can be
addressed by an
iterative algorithm \autoref{alg:WassBary}, where each iteration
involves two steps. First, an \emph{assignment step} computes the
optimal assignment given the $W_q$ metric between each input diagram and the
current barycenter estimation. Second, in the \emph{update step},
the Fréchet energy
is minimized by computing, for each barycenter point, its \emph{ground
barycenter} in the birth/death plane.
This is achieved
by updating each barycenter point to its optimal location, given the
assignments computed in the previous step.}

\begin{figure}
\centering
 \includegraphics[width=\columnwidth]{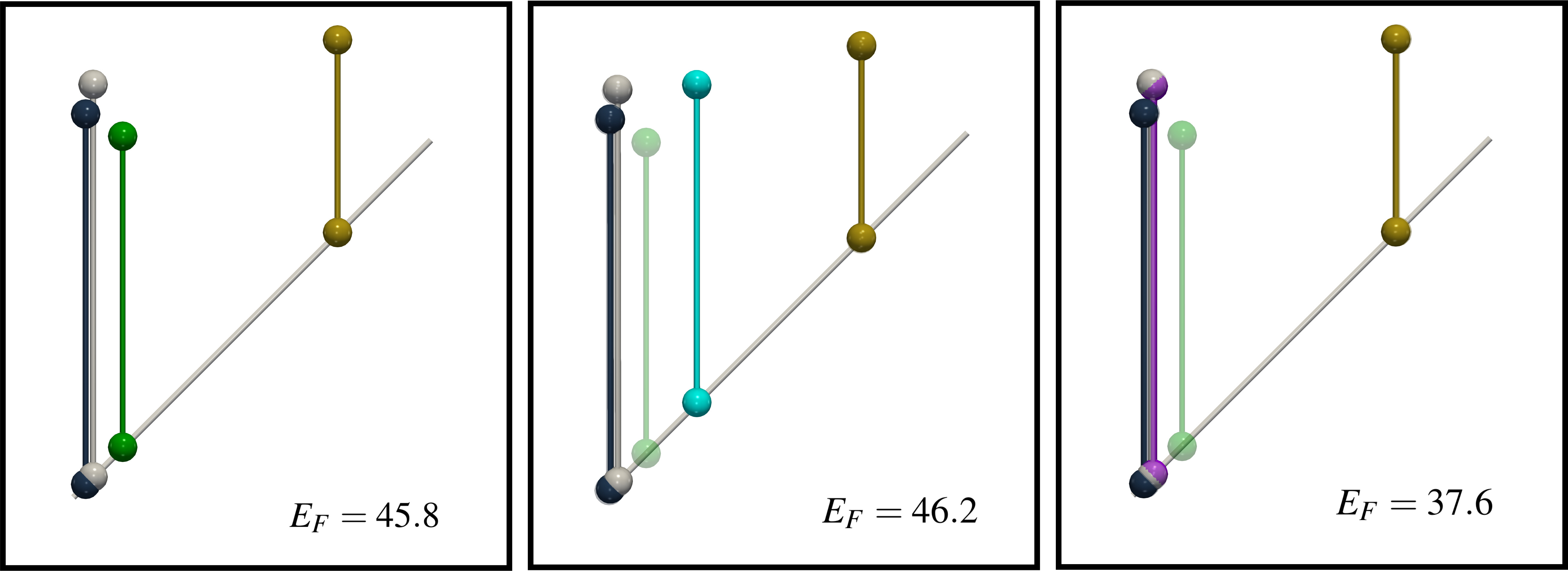}
\caption{{Simple example where computing the arithmetic mean instead of
optimizing $\br_{{q}}$ increases the Fréchet energy {(noted
$E_F$)}
{for $q = 1$.}
We have
three simple persistence diagrams,
in {dark} blue, gray and
yellow, each having a single point. For this problem, the transport plans are
fixed and the barycenter has only one point. On the left we initialized the
barycenter as the diagram encoded in green. In the middle, we have the candidate
of the barycenter encoded in cyan when computing an arithmetic mean after one
iteration{. We}
can see that the Fréchet energy {(for $q = 1$)} increased. On the
right, we have a candidate for the barycenter encoded in purple when optimizing
$\br_{{q}}$ instead, this time displaying a decrease of the Fréchet
energy at one iteration.}}
\label{fig:Showing}
\end{figure}

{For $q = 2$, the ground barycenter can be simply obtained by computing,
for each barycenter point,
the arithmetic means of its assigned points in the input diagrams
\cite{Turner2014, vidal20}. However, for $q \neq 2$, such a simple update
procedure
cannot be considered. As illustrated in \autoref{fig:Showing}, an update based
on the arithmetic mean may increase the Fréchet energy for $q \neq 2$, hence
potentially preventing \autoref{alg:WassBary} from converging toward a
satisfying result.}

Instead, to generalize
{the computation of ground barycenters for $q \neq 2$,}
we
{consider}
the
following function, representing the ground barycenter
{in the birth/death plane:}
{
\begin{equation}
\label{eq_groundBarycenter}
\br_{{q}}:\begin{cases}
\R^2\times\hdots\times\R^2 \to \R^2\\
(y_1,\hdots,y_m) \mapsto \displaystyle\argmin_{{x\in\R^2}} \sum\limits_{i=1}^m \lambda_i c_{q}(x,y_i)
\end{cases}.
\end{equation}}

{To optimize \autoref{eq_groundBarycenter}, 
we use gradient descent, by leveraging the automatic differentiation 
capabilities of PyTorch {(for which we give the parameters used in Appendix A)}.
This optimization procedure is plugged into \autoref{alg:WassBary} in the \emph{update step} (ground barycenter
computation line).}
{In practice we give a maximum number $T$ of overall iterations.}
{Regarding the time complexity of \autoref{alg:WassBary},
computing an optimal matching between two persistence diagrams using
traditional linear programming methods requires a runtime complexity of
$\mathcal{O}(K^3)$ \cite{peyre2019computational}. Also,
{to update the barycenter,}
a gradient descent
scheme to compute $\br_q$ has a complexity of
$\mathcal{O}(mT')$\cite{nocedal2006numerical} with $T'\leq 250$ the maximum
number of iterations {for the optimization of
\autoref{eq_groundBarycenter}}. Thus, overall \autoref{alg:WassBary} has a
runtime complexity of $\mathcal{O}\left(T(mK^3 + mT'K)\right)$ in practice.}
We noticed that taking $T < 10$ is sufficient for convergence. Assumptions for
achieving convergence are discussed in the next section.

\subsection{{Convergence}}
\label{sec_genericGroundBarycenters}

\begin{figure*}[tb]
\centering
 \includegraphics[width=\linewidth]{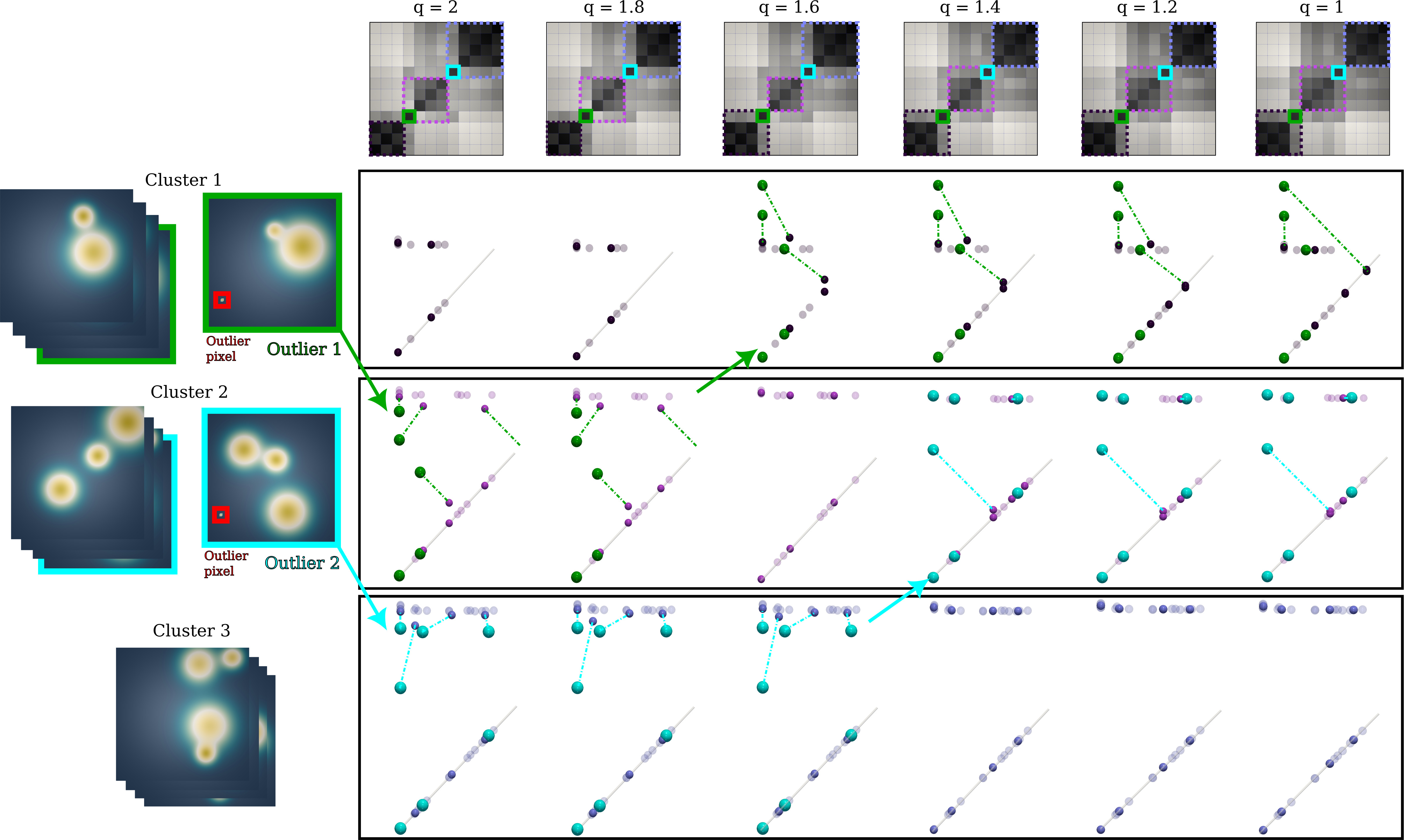}
 \caption{Comparison of clustering results on an ensemble of diagrams of
Gaussian mixtures. On the left we have the 3 clusters: one cluster of 2
Gaussians {(top)}, one cluster of 3 Gaussians {(middle)} and one
cluster of 4 Gaussians {(bottom)}. In the first and second clusters, we
inserted an outlier (highlighted in green and cyan respectively) by
{setting one isolated pixel {(highlighted in red)} to an arbitrarily high value.}
Those pixels result in persistent pairs in the
corresponding diagrams. On top we have the distance matrices of
$W_{{q}}$ for $q \in \{2, 1.8, 1.6, 1.4, 1.2, 1\}${.
In the distance matrices, the
clustering results are shown with dashed squares
(clusters are colored in dark purple, purple and pale purple) while the
outlier diagrams are indicated with a plain square (green and cyan).}
In the three
{frames;}
we
visualize the evolution of each cluster and their barycenters for each $q$. Each
{frame}
corresponds to a cluster
{(top: cluster 1, middle: cluster 2, bottom: cluster 3).
The outlier diagrams are colored in green and cyan. The barycenters are shown
in opaque while the diagrams of each cluster are shown in transparent.}
{We observe} that for $q \in \{2,1.8\}$ the green outlier is
{incorrectly} assigned to the second cluster (as it exhibits the same
number of persistence pairs, 3, as the entries of cluster 2),
{then, from $q = 1.6$, it is {correctly} assigned
to the first cluster as shown by the green arrow (it shifts from the second
frame to the first one). Similarly,
given its number of persistence pairs, the cyan outlier is incorrectly assigned
to the third cluster until $q=1.6$, then from $q = 1.4$ it shifts to the second cluster as indicated by the cyan arrow.}} 
 \label{fig:GaussToy}
\end{figure*}

{Tanguy et al. 
\cite{tanguy2025computingbarycentresmeasuresgeneric}
prove,
in the setup of probability
measures,
that 
a fixed-point method
for minimizing \autoref{eq_groundBarycenter}
converged under certain assumptions
\cite{tanguy2025computingbarycentresmeasuresgeneric}. In this section, we
review these assumptions in the setup of persistence diagrams to argue the
convergence of our overall approach.}

\noindent \begin{Hyp}
For all $(y_1,\hdots,y_m)\in \R^2\times\hdots\times\R^2$, for all $\lambda_1,\hdots,\lambda_m$ barycentric coefficients, $\displaystyle \argmin_{{x\in\R^2}} \sum\limits_{i=1}^m \lambda_i c_{q}(x,y_i)$ is reduced to a single element.
\end{Hyp}
We {prove that this assumption is satisfied in our case} in
Appendix {B}, { when $q > 1$}.
{When $q = 1$, this assumption does not hold in general
{(due to the presence of collinear points, in particular on the
diagonal)}.}
{In practice, this may lead to numerical instabilities when considering
$q = 1$, especially when ground barycenters are not unique.}
{Thus,
{our approach does not support the case $q=1$ in general and}
we recommend to take $q>1$ (unless the {input} diagrams are
simple {and not prone to numerical instabilities, e.g.
\autoref{fig:GaussToy})}.}

Under those assumption{s,}
and denoting {the
Fréchet energy} $E_F(B) = \sum\limits_{i=1}^m \lambda_i W^q_{q}(B,X_i)$,
{Tanguy et al.}
\cite{tanguy2025computingbarycentresmeasuresgeneric} {show that for two consecutive iterates $B^{(t)}$ and $B^{(t+1)}$}, we have $E_F(B^{(t)}) \geq
E_F(B^{(t+1)})$. This means that fixed-point iterations $(B^{(t)})$ decrease the
energy optimized in
{\autoref{eq:2Bary}.} {However, while the iterates
decrease $E_F$, it is not guaranteed that the
{obtained}
fixed-point
is a global
(even local) minimum of $E_F$}. {For $q \in (1,2)$,} a resulting fixed
point is a barycenter that is more
robust to the presence of an outlier, in the initial set $X_1,\hdots,X_m$,
compared to a $W_2$ barycenter. \autoref{fig:BaryComparison} illustrates
{this}
difference when computing a barycenter with
an outlier. {Basically, the closer $q$ is to 2, the more it
behaves
{like a \emph{mean}.}
Conversely, the closer $q$ is to 1, the more it behaves like a
{\emph{median} (i.e., describing typical central values, less
sensitive to outliers than averages).}} 

\begin{figure}[tb]
\centering
 \includegraphics[width=\columnwidth]{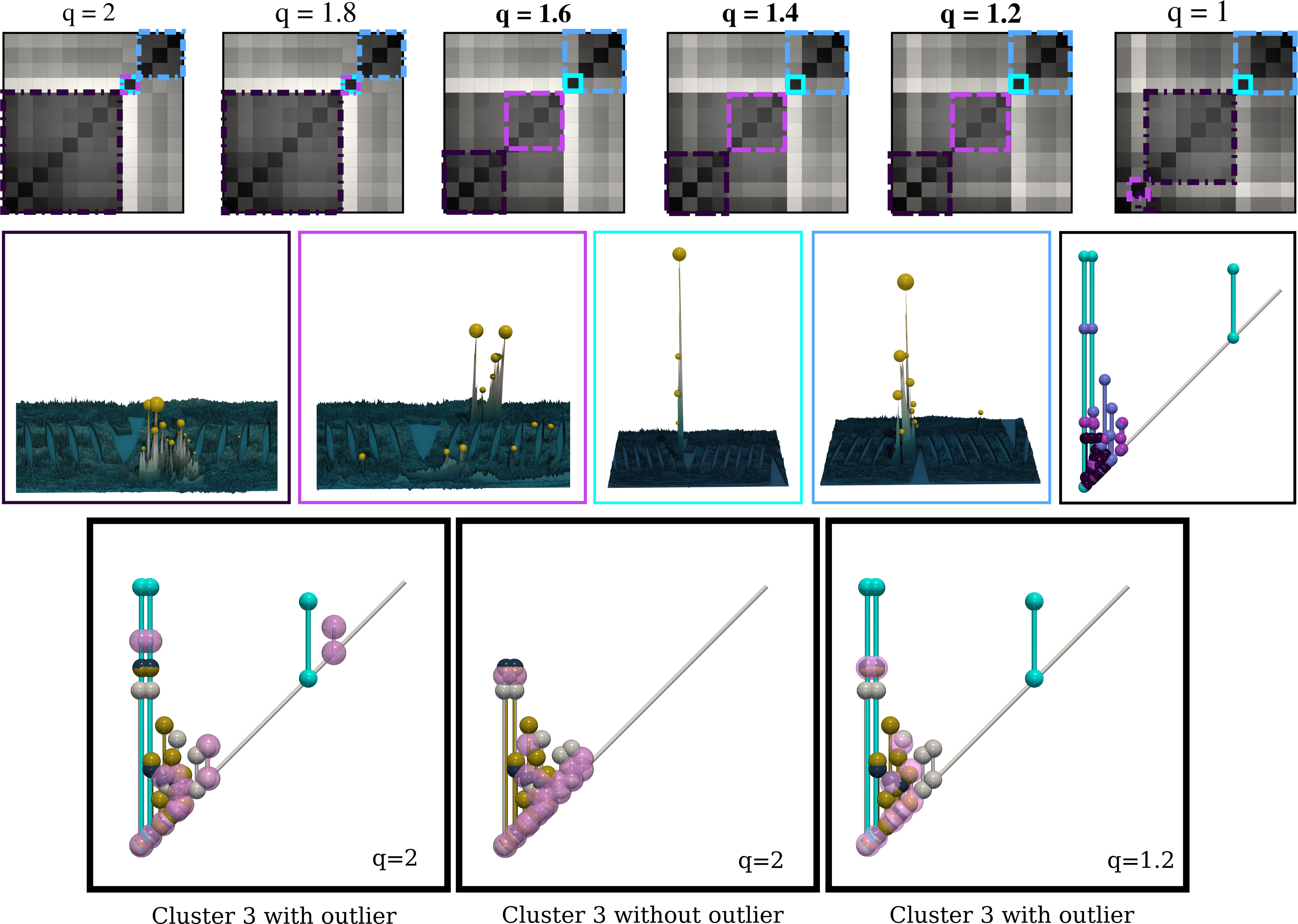}
 \caption{Visual comparison of distance matrices using $W_{{q}}$ for $q
\in \{2, 1.8, 1.6, 1.4, 1.2, 1\}$ on the \textit{Volcanic Eruption} ensemble
and the clustering results.
This ensemble of 12 persistence diagrams has a
natural outlier highlighted in cyan on the distance matrices. On the top, we can
see that for $q \in \{2, 1.8\}$, the 
{clustering}
algorithm keeps the outlier
alone, groups the 8 first diagrams together and groups the last three together.
Then starting from 1.6 to 1.2, the correct clusters are returned. But for $q =
1$, we can see that the clusters are not discriminated enough. In the middle we
have one representative scalar field for each cluster, and on the right
the corresponding diagrams, the cyan scalar field and diagram being the outlier. 
{On the bottom, we have a visual comparison of three barycenters (pink) of the last 
cluster in three cases: one computed with the outlier in cyan when $q = 2$, one without the outlier when $q = 2$ and one with the outlier when $q = 1.2$. We witness the influence of the outlier on the barycenter on the left, as 
there are two {persistent} pairs
{(in pink)}
higher than the ones in the other three diagrams of the cluster. Also, we
notice the presence of an isolated
pair above the diagonal that is 
generated by the isolated cyan
pair. In the other
cases, the
barycenters {(pink)} are very
similar, showing the robustness of this $1.2$-barycenter to the presence of this
outlier.}}
 \label{fig:Volcano}
\end{figure}

\section{Results} \label{sec:result}
\noindent

{This section presents}
two applications of
{our}
robust
barycenter{s} (\autoref{sec:RobustBary})
{along with detailed experiments.}
The experimental results are obtained on a computer with {an}
NVIDIA Geforce RTX 2060 (Mobile Q) with 6 GB of dedicated VRAM.
{Our}
methods
{were}
implemented on Python, using Pytorch for computations on the GPU.
{Given the specification of our GPU, we were memory-limited in
our experiments, given the modest dedicated VRAM. This led us to run
experiments with diagrams thresholded by persistence, to reach a typical size of
$\sim$20 points per diagram (as a reminder, in practice, all diagrams need to
be augmented, \autoref{sec:PDs}).}
We ran some experiments on two public ensembles \cite{ensembleBenchmark}
described in \cite{pont_vis21}. One is an acquired 2D ensemble and the other is
a simulated 3D ensemble, both
{selected}
from past SciVis contests
\cite{scivisOverall}. For the experiments, only the persistence pairs containing
maxima were considered.

\subsection{Clustering  on the persistence diagram metric space}
\label{sec:kmeans}

The first {natural} application
{of our robust barycenters consists in using them for
the problem of clustering an ensemble of persistence diagrams}
$X_1,\hdots,X_N$. In particular,
clustering methods on ensembles of persistence diagrams group together subset of
members that have similar topological structures, highlighting trends of
topological features in the
ensemble.

For this{,} we consider the classic clustering method, the $k$-means
algorithm.
{This}
is an iterative algorithm alternating between two phases:
\textit{computing k barycenters} and \textit{labeling the elements into k
clusters}.
At first, $k$ cluster barycenters $B_j$ with $j\in \{1,\hdots,k\}$ are
initialized as $k$ diagrams in the initial input set $X_1,\hdots,X_N${,
typically} using the $k$-means++ method \cite{celebi13}. Then, the labeling
phase consists in assigning each diagrams $X_n$ to the closest barycenter $B_j$
by using the $W_{{q}}$ distance. After the labeling phase, the
barycenters are updated by computing new barycenters based on the new $k$
clusters using \autoref{alg:WassBary}. The algorithm stops when reaching a
maximum number of iterations or when converging (i.e{.,} when the labels
do not change anymore).
However, when using $W_2$ as a distance, the presence of an outlier in an ensemble of persistence diagrams can incorrectly influence the barycenters in the $k$-means algorithm, and as a consequence the output labels of the clustering. 

We leverage the robustness of the $W_{q}$ barycenters for clustering problems
when there are outliers in the ensemble to be clustered. We show a case in
\autoref{fig:GaussToy} where we artificially injected outlier {pixels}
in an ensemble of
{synthetic}
scalar fields {(a common degradation in real-life noisy data)}.
{This results in outlier diagrams in the input ensemble. We illustrate}
the clustering results with different $q\in \{2, 1.8, 1.6, 1.4, 1.2,
1\}$. This experiment show{s} that for $q$ lower than $1.6${,}
the
{resulting}
clustering put together the outliers into the correct groups, while
for $q = 2$ the outliers are incorrectly clustered. This can also be seen in
\autoref{fig:Volcano} where an outlier is naturally present in the ensemble of
scalar fields.
{In this example, our generic barycenters enable the computation of the
correct clustering, as off $q = 1.6$. Moreover, as illustrated on the right of
\autoref{fig:Volcano} for the last cluster, our generic barycenters are more
representative, visually, of the input diagrams as the importance of outlier
persistence pairs is decreased in our framework.}

\begin{figure}[tb]
\centering
 \includegraphics[width=\columnwidth]{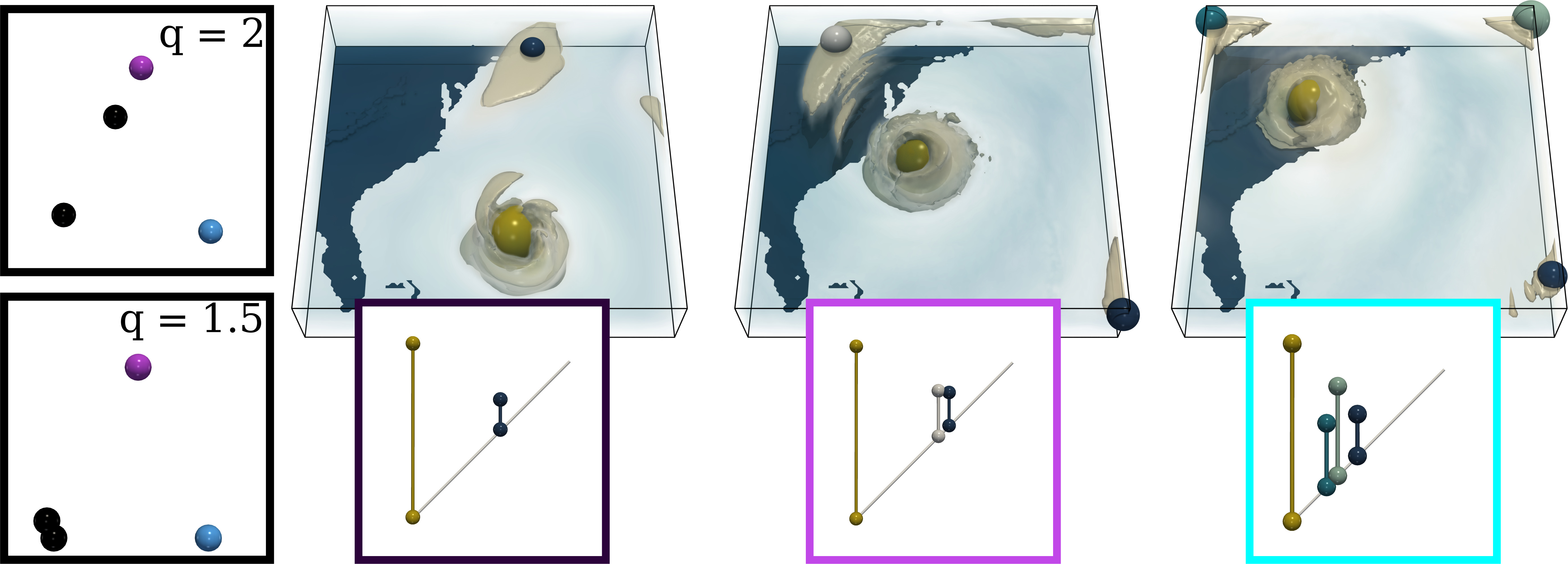}
 \caption{{Visual comparison of 2D planar layouts of Wasserstein
barycenters generated by the Wasserstein dictionary encoding method with $q=2$
(top-left) and $q=1.5$ (bottom-left). {The initial \textit{Isabel}
ensemble is divided in 3
{classes}
of 4 diagrams, and we removed 4 entries to create an imbalance in cluster
sizes.}
{We show {a} representative scalar field and
diagram for each cluster highlighted in three different colors.}
{In}
the planar layouts, the points, representing the barycenters, are colored by
{their}
ground truth
{classification}.
For $q = 2$, {the}
method
{misplaces}
a point from the first cluster
(dark purple) {near the
purple points.}
For $q = 1.5$, the method provides well separated classes,
thus
{yielding a planar projection that is more faithful to the
ground-truth classification.}}}
 \label{fig:Isabel}
\end{figure}

\subsection{Wasserstein dictionary encoding of persistence diagrams} \label{sec:PDDict}

{Another application consists in using our robust barycenters as a core
procedure for dictionary based encodings of ensembles of persistence diagrams
\cite{sisouk2024}.}
Let $X_1,\hdots,X_N$ be an ensemble of persistence diagrams.
{A Wasserstein dictionary encoding aims at optimizing}
a set of
persistence diagrams $\Dict^* = \{a_1^*,\hdots,a_m^*\}$ {(}called
dictionary{)} and $N$ vectors of barycentric coefficients $\Lambda^* =
\{\Lamb_1^*,\hdots,\Lamb_N^*\}${$\in\Sigma_m^N$} by solving:
\begin{equation}\label{eq:WassDictionary2}
\argmin_{{\Dict\in\mathbb{D}^m, \Lambda\in\Sigma_m^N}} \displaystyle\sum\limits_{\ell = 1}^N W_2^2\big(B_2(\Dict, \Lamb_{\ell}), X_{\ell}\big),
\end{equation}
where $B_2(\Dict, \Lamb_{\ell})$ denotes a $W_2$ barycenter of $\Dict$
{under}
barycentric coefficients $\Lamb_l$.
Informally, this framework works as a lossy compression
for persistence diagrams. The goal is
{to optimize}
a smaller set of
persistence diagrams ($m \ll N$) and $N$ vectors of barycentric weights such
that {the $N$ Wasserstein barycenters defined by the barycentric weights
are good approximations of the $N$ input diagrams. This results in an encoding
of much smaller size as only the dictionary and the $N$ barycentric weights
need to be stored to disk.}
This
{framework}
has two
main applications: data reduction and dimensionality reduction. Naturally this
framework can be extended to other Wasserstein distances. {Then,}
\autoref{eq:WassDictionary2} becomes{:}
\begin{equation}\label{eq:WassDictionaryPQ}
\argmin_{{\Dict\in\mathbb{D}^m, \Lambda\in\Sigma_m^N}} \displaystyle\sum\limits_{l = 1}^N W_{q}^q\big(B_{q}(\Dict, \Lamb_l), X_l\big),
\end{equation}
where $B_{q}(\Dict, \Lamb_l)$ denotes a $W_{q}$ barycenter returned by
\autoref{alg:WassBary}.
{For $q=2$, Sisouk et al. \cite{sisouk2024} introduced the analytic
expression of the gradient $\nabla B_2(\Dict, \Lamb_l)$ with respect to
$a_1,\hdots,a_m$ and $\Lamb_l$, enabling a simple gradient descent scheme for
the optimization of \autoref{eq:WassDictionaryPQ}. However, for $q\neq2$, since
ground barycenters are no longer obtained as arithmetic means, but by an
interative, fixed-point method (\autoref{sec_genericGroundBarycenters}), the
gradient of the energy associated with \autoref{eq:WassDictionaryPQ} cannot be
derived analytically. Instead, we rely on automatic differentiation (implemented
in PyTorch) and use Adam \cite{kingma2017adammethodstochasticoptimization} to
optimize both the dictionary $\Dict = \{a_1,\hdots,a_m\}$ and the vectors of
barycentric coefficients $\Lambda = \{\Lamb_1,\hdots,\Lamb_N\}$.}

This extension results in a Wasserstein dictionary method that is more stable to the presence of outliers in the original input ensemble. Moreover, this extension lets us improve the initialization method for the dictionary. At first, the dictionary was chosen as the $m$ elements that are the farthest to each other in the initial set \cite{sisouk2024}. But extending \autoref{eq:WassDictionary2} to \autoref{eq:WassDictionaryPQ} allowed the dictionary to be initialized as barycenters issued from a $k$-means algorithm with $k = m$. From our experience, this initialization results in a better optimized energy (\autoref{eq:WassDictionaryPQ}) and stability to the presence of outliers compared to the initial one \cite{sisouk2024}. 
We showcase this extension by applying a Wasserstein dictionary method using
$W_{q}$. \autoref{fig:Isabel} shows a comparison of 2D planar layout of the
barycenters, generated by the dictionaries and vectors optimized with
\autoref{eq:WassDictionary2} and \autoref{eq:WassDictionaryPQ}, on the
\textit{Isabel} ensemble where we removed some entries to artificially
inject an outlier {(see the caption of \autoref{fig:Isabel} for more
details)}.
{Specifically,}
this 2D planar layout is a direct application of the dictionary encoding when
the dictionary has three atoms in it. After optimizing the dictionary, with the
three Wasserstein distances between them, the cosine law can be used to form a
triangle in $\R^2$ and then use the vectors as barycentric coefficients in
$\R^2$.
This experiments shows that this extension is more stable to the presence of
outliers,
{resulting in planar projections that are more faithful to the ground
truth classification.}

\subsection{Computation time comparison}

\begin{table}[h!]
\caption{Running times (in seconds) {for
computing a {$W_q$} barycenter}.}
\centering
{
\begin{tabular}{ |p{2cm}|r|r|r|r|r|  }
 \hline
 Method &$m = 4$ & $m = 6$ & $m = 8$ & $m = 10$ & $m = 12$ \\
 \hline
 Arithmetic Mean & {0.1} & {0.3} & {4.5} & {36.7} & {100.1} \\
 \hline 
{$\br_{{2}}$ (Pytorch)} & {5.4} & {5.0} & {9.7} & {47.3} & {147.7}\\
\hline
  {$\br_{ {1.8}}$ (Pytorch)} & {5.1} & {5.0} & {9.9} & {50.1}& {135.5}\\
 \hline
  {$\br_{ {1.6}}$ (Pytorch)} & {5.0} & {5.4} & {9.4} & {54.1}& {149.2}\\
 \hline
   {$\br_{ {1.4}}$ (Pytorch)} & {4.4} & {6.5} & {10.6} & {56.0}& {147.6}\\
 \hline
   {$\br_{ {1.2}}$ (Pytorch)} & {5.2} & {7.6} & {11.6} & {57.6}& {152.7}\\
 \hline
\end{tabular}}
\label{Tab:Speed}
\end{table}
{In this {section}, we compare the time needed to compute a
barycenter{, with regard to the $W_2$ metric,} using the arithmetic mean
between points of $\R^2$, and the time when computing
{$\br_{2}$}.
Our
experiment consists in computing a barycenter of
persistence diagrams {for} the Isabel ensemble (\autoref{sec:WassDist}),
{where each diagram is thresholded by persistence to feature only {$\sim
20$ pairs and where $T$ is set to $5$ for this experiment}. We}
report the {resulting} running times in \autoref{Tab:Speed}.
{When considering $m=4,6$
input diagrams, the computation based
on the arithmetic mean is 50 times faster than the one based on
$\br_{{2}}$. However, for a larger ensemble ($m=8,10,12$), the
difference is reduced.
{This
is because
diagrams need to be
augmented prior to computing matchings between them, and that, consequently,
the size of the barycenter increases significantly when $m$ increases, yielding
an assignment time (cubic complexity) being prevalent over the update time
(see \autoref{alg:WassBary} and \autoref{sec_optimization}).}}

\section{Conclusion}

In this paper, we showcased the utility of a method for computing a robust
Wasserstein barycenters of persistence diagrams. Specifically, we adapted a
recent fixed-point method algorithm
\cite{tanguy2025computingbarycentresmeasuresgeneric} to the case of persistence
diagrams. We first gave a reminder on this fixed point method framework to
compute this robust barycenter{, and then gave insights on the
choice of $q$ depending on the characteristic one wants from the barycenter.} We
also gave a formal proof of the necessary hypothesis for the convergence of this
method in the Appendix. Then we presented two applications of this robust
barycenter to clustering and dictionary encoding of persistence diagrams in the
presence of outliers. {We also noticed that, across all
{our}
experiments,
taking $q\in[1.2,1.4]$ yields the best results.}
{We believe that our work on the robustness of Wasserstein barycenters
is a useful step toward improving their applicability
in
the
analysis of real-life ensembles of persistence diagrams.}

A legitimate direction for future work is the generalization of such robust
barycenters to other topological descriptors such {as}
merge trees{, for which barycentric frameworks derived from the
Wasserstein distance have been proposed} \cite{pont_vis21}.
{Also, we}
will
continue our investigation of the adaption of
methods from Optimal
Transport to ensembles of topological representations, as we believe it can
become a key solution in the long term for the advanced analysis of large-scale
ensemble {datasets}.

\section*{Acknowledgments}{
\small
This work is partially supported by the European Commission grant
ERC-2019-COG \emph{``TORI''} (ref. 863464, \url{https://erc-tori.github.io/}). JD acknowledge the support of the “France 2030” funding ANR-23-PEIA-0004 (“PDE-AI”) and the support of the funding SOCOT - ANR-23-CE40-0017.}

\bibliographystyle{abbrv}

\bibliography{bibli}

\vspace{-4ex}
\begin{IEEEbiography}[
{\includegraphics[width=1in,height=1.25in,clip,
keepaspectratio]{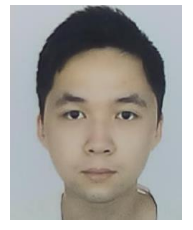}}]{Keanu Sisouk}
is a Ph.D. student at Sorbonne University. He received his master's degree in Mathematics from Sorbonne University in 2021. His fields of interests lie on topological methods for data analysis, optimal transport, optimization methods, statistics and partial derivative equations.
\end{IEEEbiography}
\vspace{-4ex}
\begin{IEEEbiography}[
{\includegraphics[width=1in,height=1.25in,clip,
keepaspectratio]{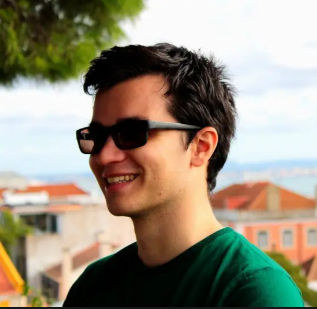}}]{Eloi Tanguy}
is a Ph.D. student at Université Paris-Cité. he received his master's degree in
Mathematics from ENS Paris-Saclay in 2022 after his studies at Ecole
Polytechnique. His research interests are primarily in optimal transport, in
particular OT Barycentres, the Sliced Wasserstein Distance
and GMM-OT.
\end{IEEEbiography}
\vspace{-4ex}
\begin{IEEEbiography}[
{\includegraphics[width=1in,height=1.25in,clip,
keepaspectratio]{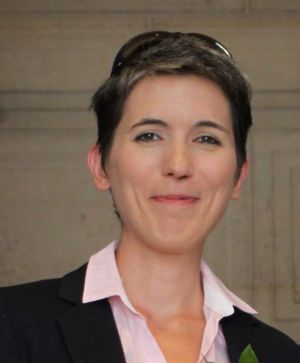}}]{Julie Delon}
received the Ph.D. degree in Mathematics from the Ecole Normale Supérieure Cachan in 2004.
She is currently a professor at ENS Paris since 2025. She was a CNRS researcher affiliated with TELECOM ParisTech and obtained her professor tenure at Paris-Cité University in 2013.
Her research interests lies in optimal transport, image processing, inverse problems and stochastic models for image restoration and editing.
\end{IEEEbiography}
\vspace{-4ex}
\begin{IEEEbiography}[
{\includegraphics[width=1in,height=1.25in,clip,
keepaspectratio]{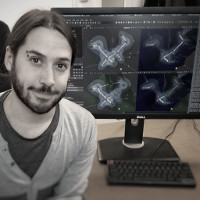}}]{Julien Tierny}
received the Ph.D. degree in Computer Science from the University
of Lille in 2008.
He is currently a CNRS research director, affiliated with
Sorbonne University.
His research expertise lies in topological methods for data analysis
and visualization.
He is the founder and lead developer of the Topology ToolKit
(TTK), an open source library for topological data analysis.
\end{IEEEbiography}

\appendices

\section{Parameters for Adam Optimizer}

{Regarding the optimizer used to update the barycenter in the
algorithm, we used the following parameters: a learning rate $\ell r = 1$, and
the optimization stops if either the maximum number of iterations $T' = 250$ is
reached or two iterates of the energy are close (under a threshold of
$10^{-15}$).}

\section{Sufficient Conditions for Assumption 1}

\begin{Cond}
    Let $y_1, \hdots, y_m \in \R^d$ and $(\lambda_1, \hdots, \lambda_m) \in (0,
    1)^m$ such that $\sum_i \lambda_i = 1$. Then for $q \in (1, +\infty)$,
    the function defined as:
    \[
    V_q := x \longmapsto \sum_{i=1}^m \lambda_i \|x-y_i\|_2^q,
    \]
    has a unique minimiser in $\R^d$. If $m\geq 3$ and there exists $i_1 < i_2 <
    i_3 \in \{1, \hdots, m\}$ such that the points $y_{i_1}, y_{i_2}$ and
    $y_{i_3}$ are not on a common affine line, then $V_1$ also has a unique
    minimiser.
\end{Cond}

\begin{proof}
    \step{1}{Case $q>1$.}

    For $q>1$ and a fixed $y\in \R^d$, introduce the function $h_q := x
    \longmapsto \|x-y\|_2^q$. We begin by showing that $h_q$ is strictly convex.
    Take $x_1, x_2 \in \R^d$ and $t\in (0, 1)$. We re-write:
    \begin{align*}
        h_q(tx_1 + (1-t)x_2) = \|t(x_1 - y) + (1-t)(x_2-y)\|_2^q.
    \end{align*}
    Introduce $u := x_1 - y$ and $v := x_2 - y$. By convexity of $\|\cdot\|_2$,
    we have $\|tu + (1-t)v\|_2 \leq t\|u\|_2 + (1-t)\|v\|_2$. We consider the
    equality and strict inequality cases separately.
    \begin{enumerate}
        \item If $\|tu + (1-t)v\|_2 < t\|u\|_2 + (1-t)\|v\|_2$, we use
        consecutively the fact that $a \longmapsto a^q$ is increasing and convex
        on $\R_+$:
        \begin{align*}
            \|tu + (1-t)v\|_2^q &< (t\|u\|_2 + (1-t)\|v\|_2)^q \\
            &\leq t\|u\|_2^q + (1-t)\|v\|_2^q.
        \end{align*}

        \item If $\|tu + (1-t)v\|_2 = t\|u\|_2 + (1-t)\|v\|_2$, equality in the
        triangle inequality for $\|\cdot\|_2$ yields that $u$ and $v$ are
        positively co-linear, leading to the two following alternatives:
        \begin{enumerate}
            \item If $v = 0$ then we show the strict inequality as follows:
            \begin{align*}
                \|tu\|_2^q = t^q\|u\|_2^q < t\|u\|_2^q,
            \end{align*}
            where the inequality comes from the fact that $q>1$ and $t\in (0,
            1)$, with $u\neq 0$ (indeed, if $u=0$ then we have $u=v=0$ yielding
            $x_1=x_2$ which is a contradiction).
            \item If $v\neq 0$ then there exists $\alpha \geq 0$ such that
            $u=\alpha v$. This implies that $\|u\|_2\neq \|v\|_2$: if equality
            held, then since $\alpha \geq 0$ we obtain $\alpha = 1$, then $u=v$
            yields $x_1=x_2$ which is a contradiction. Since $\|u\|_2\neq
            \|v\|_2$, we can apply the strict convexity of $a \longmapsto a^q$
            on $\R_+$, which shows:
            \begin{align*}
                \|tu + (1-t)v\|_2^q &= (t\|u\|_2 + (1-t)\|v\|_2)^q \\
                &< t\|u\|_2^q + (1-t)\|v\|_2^q.
            \end{align*}
        \end{enumerate} 
    \end{enumerate}
    In all cases, we obtain the inequality:
    \[
    h_q(tx_1 + (1-t)x_2) < th_q(x_1) + (1-t)h_q(x_2),
    \]
    showing strict convexity of $h_q$. As a convex combination of strictly
    convex functions, $V_q$ is strictly convex. Since $V_q$ is coercive, we
    conclude that it admits a unique minimiser.

    \step{2}{Case $q=1$.}

    We now assume that $m\geq 3$ and that there exists $i_1 < i_2 < i_3 \in \{1,
    \hdots, m\}$ such that $y_{j_1}, y_{j_2}$ and $y_{i_3}$ are not on a common
    affine line. We prove that $V_1$ is strictly convex: let $x_1 \neq
    x_2\in\R^d$ and $t\in (0,1)$, by the triangle inequality (similarly to the
    case $q>1$):
    \begin{align*}
        V_1\big(tx_1 + (1-t)x_2\big) 
        \leq \sum_{i=1}^m \lambda_i\big(\|t(x_1-y_i)\|_2 
        + \|(1-t)(x_2-y_i)\|_2\big),
    \end{align*}
    using the triangle inequality. Our objective is to show that in this case,
    the inequality is strict. There is equality if and only if for each $i \in
    \{1, \hdots, m\}$, $x_2-y_i = 0$ or there exists $\alpha_i \in \R_+$ such
    that $x_1-y_i = \alpha_i(x_2 - y_i)$. We now reason by contradiction and
    assume that equality holds. We distinguish two cases concerning the points
    $(y_{i_k})_{k=1}^3$ from the assumption.
    \begin{enumerate}
        \item If there exists $k \in \{1, 2, 3\}$ such that $x_2 - y_{i_k} =
        0$, without any loss of generality we take $y_{i_1} = x_2$, then
        $y_{i_1} = x_2$, furthermore the assumption on $(y_{i_k})_{k=1}^3$
        implies that:
        \[
        \forall i \in \{i_2, i_3\},\; x_2 - y_i \neq 0.
        \]
        Using these properties, we deduce from the equality in the triangle
        inequality that there exists $\alpha_{i_k} \geq 0$ such that $x_1 -
        y_{i_k} = \alpha_{i_k}(x_2 - y_{i_k})$ for $k\in\{2,3\}$. Note that the
        $\alpha_{i_k} \neq 1$, otherwise $x_1 - y_{i_k} = x_2 - y_{i_k}$ which
        is impossible since $x_1 \neq x_2$. We can now re-rewrite the equality
        as
        \[
        y_{i_k} = x_1 + \cfrac{\alpha_{i_k}}{1 - \alpha_{i_k}}(x_1 - x_2),
        \]  
        concluding that $y_{i_2}$, $y_{i_3}$ and $x_2$ are on the common affine
        line $x_1 + \R(x_1 - x_2)$, which is a contradiction as $x_2 =y_{i_1}$.
        We conclude that the equality cannot hold in the triangle inequality in
        this case.

        \item If $\forall k \in \{1, 2, 3\},\; x_2 - y_{i_k} \neq 0$, then the
        triangle equality condition provides the existence of $\alpha_{i_k} \geq
        0$ such that $x_1 - y_{i_k} = \alpha_{j_k}(x_2 - y_{i_k})$. Again
        $\alpha_{i_k} \neq 1$ for the same reasons discussed above. Similarly we
        re-write the equality as:
        \[
        y_{i_k} = x_1 + \cfrac{\alpha_{i_k}}{1 - \alpha_{i_k}}(x_1 - x_2),
        \]
        concluding that the three points $(y_{i_k})_{k=1}^3$ are on the common
        affine line $x_1 + \R (x_1 - x_2)$ which is a contradiction, hence
        equality cannot hold in the triangle inequality. 
    \end{enumerate}
 
    In both cases, we have proven that equality in the triangle inequality
    cannot happen, yielding:
    \begin{align*}
        V_1\big(tx + (1-t)x_2\big) <tV_1(x) + (1-t)V_1(x_2),
    \end{align*}
    which shows the strict convexity of $V_1$. Since $V_1$ is coercive, we can
    conclude that it admits a unique minimiser.
\end{proof}

\begin{Rk}
    When $q = 1$ it is crucial that at least three points are not on a common affine line. For instance if we have only two points, and take $\lambda_1 = \lambda_2=
    1/2$, then $\argmin_{x} \|x-y_1\|_2 + \|x-y_2\|_2 = \{ty_1 + (1-t)y_2\ | \
    t\in[0,1]\}$. 
\end{Rk}

\end{document}